%% file: main.tex
\documentclass[runningheads]{llncs}
\usepackage[T1]{fontenc}
\usepackage{hyperref}
\usepackage{comment} 
\usepackage{graphicx}
\usepackage{cite}
\usepackage{booktabs}
\usepackage{amssymb}
\usepackage[misc]{ifsym}

\begin{document}

\title{An Examination of Wearable Sensors and Video Data Capture for Human Exercise Classification}
\titlerunning{Wearable Sensors and Video Data Capture for Human Exercise Classification}
 %\title{\Large Realistic Application of Wearable Sensors and Video Data Capture for Human Exercise Classification}
 
\author{Ashish Singh
\and Antonio Bevilacqua 
\and Timilehin B. Aderinola
\and Thach Le Nguyen
\and Darragh Whelan
\and Martin O'Reilly 
\and Brian Caulfield
\and Georgiana Ifrim }

\institute{Insight Centre for Data Analytics, University College Dublin, Ireland \\
\email{\{ashish.singh,antonio.bevilacqua,timi.aderinola,thach.lenguyen,b.caulfield, georgiana.ifrim\}@insight-centre.org}\\
\and Output Sports Limited, NovaUCD, Dublin, Ireland
\email{\{darragh, martin\}@ouputsports.com}\\
}

% \authorrunning{A Singh et al.}

%\author{}

\maketitle
\input{abstract}
\input{1-introduction.tex}

\input{2-related-work.tex}

\input{3-data.tex}

\input{4-methods.tex}

\input{5-experiments.tex}

\input{6-conclusion.tex}

\input{ethical_issues.tex}

\bibliography{main}
\bibliographystyle{splncs04}

\end{document}

%% file: abstract.tex
\begin{abstract}

Wearable sensors such as Inertial Measurement Units (IMUs) are often used to assess the performance of human exercise. Common approaches use handcrafted features based on domain expertise or automatically extracted features using time series analysis. Multiple sensors are required to achieve high classification accuracy, which is not very practical. These sensors require calibration and synchronization and may lead to discomfort over longer time periods. Recent work utilizing computer vision techniques has shown similar performance using video, without the need for manual feature engineering, and avoiding some pitfalls such as sensor calibration and placement on the body.
In this paper, we compare the performance of IMUs to a video-based approach for human exercise classification on two real-world datasets consisting of Military Press and Rowing exercises.
We compare the performance using a single camera that captures video in the frontal view versus using 5 IMUs placed on different parts of the body. We observe that an approach based on a single camera can outperform a single IMU by 10 percentage points on average. Additionally, a minimum of 3 IMUs are required to outperform a single camera. We observe that working with the raw data using multivariate time series classifiers outperforms traditional approaches based on handcrafted or automatically extracted features. Finally, we show that an ensemble model combining the data from a single camera with a single IMU outperforms either data modality. Our work opens up new and more realistic avenues for this application, where a  video captured using a readily available smartphone camera, combined with a single sensor, can be used for effective human exercise classification. 
\keywords {Exercise Classification \and Inertial Sensors \and Video \and Time Series Classification \and Real-World Datasets.}
\end{abstract}

%% file: 1-introduction.tex
\section{Introduction}
\label{sec:intro}
Recent years have seen an accelerated use of machine learning solutions to assess the performance of athletes.
New technologies allow easier data capture and efficient machine learning techniques enable effective measurement and feedback. In this paper, we focus on the application of human exercise classification where the task is to differentiate normal and abnormal executions for strength and conditioning (S\&C) exercises. S\&C exercises are widely used for rehabilitation, performance assessment, injury screening and resistance training in order to improve the performance of athletes \cite{Martin2018, Martin2015}. 
Approaches to data capture are either sensor-based or video-based. For sensor-based approaches, sensors such as Inertial Measurement Units (IMUs) are worn by participants \cite{Martin2015, Martin2018}. For video, a participant's motion is captured using 3D motion capture \cite{molias2017pre}, depth-capture based systems \cite{zerpa2015use}, or 2D video recordings using cameras \cite{ashishdami2022, Rahmadani}. The data obtained from these sources is processed and classified using machine learning models. 
Classification methods based on sensor data are popular in the literature and real-world applications, and yet, video-based approaches are gaining popularity \cite{ashish2020, ashishdami2022} as they show potential for providing high classification accuracy and overcoming common issues of inertial sensors.
\vspace{-0.5cm}
\begin{figure*}[ht!]
\centering
    \includegraphics[ width=1.0\textwidth]{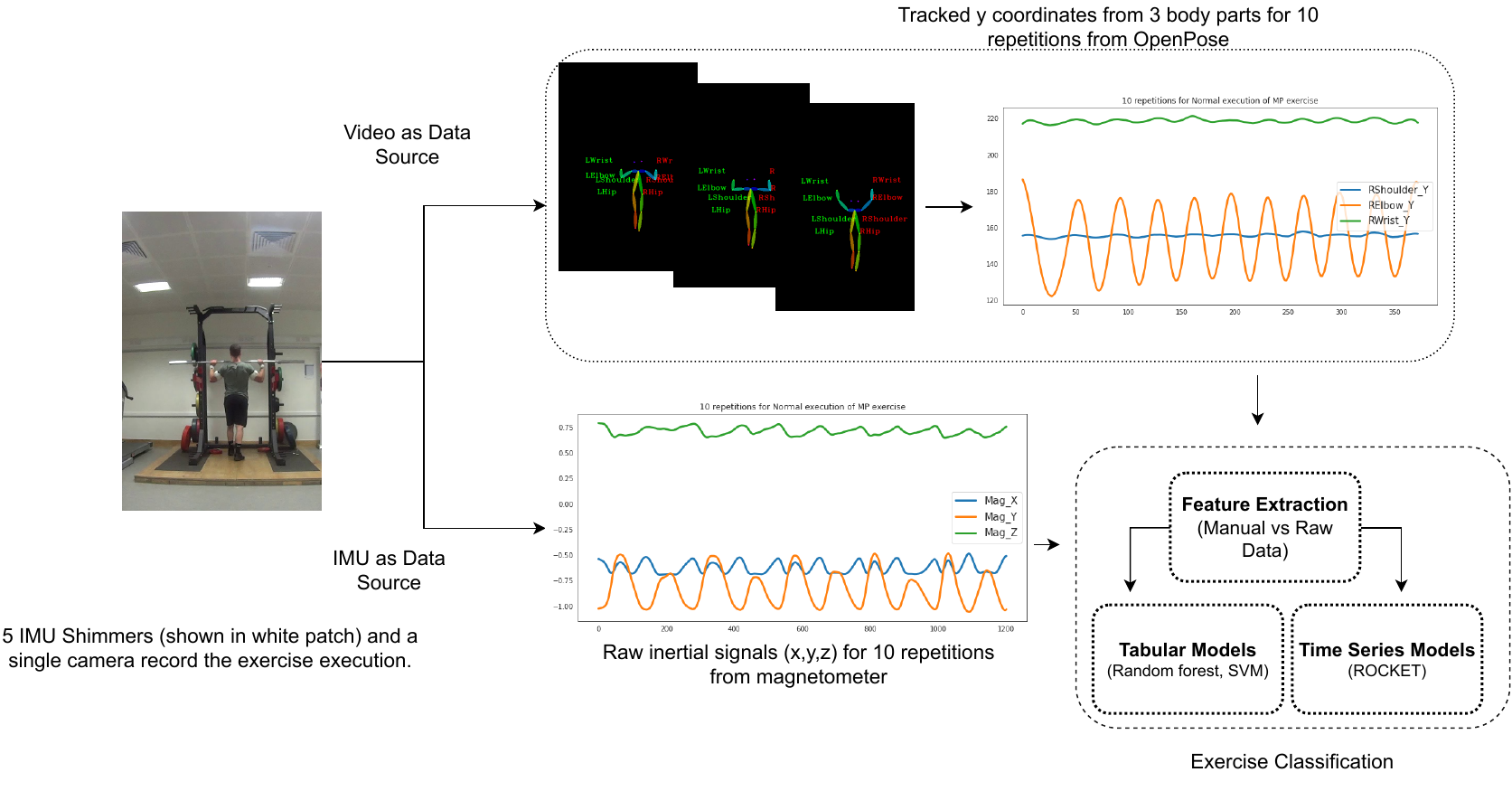}
    \caption{Comparison of video (top) and sensors (bottom) to classify human exercise movement. The upper box presents the process of obtaining multivariate data from video (only 3 out of 25 body parts shown). The bottom box shows the raw Y-signals from a single IMU placed on the participant's body (only 3 signals shown here).}
    \label{fig:overview}
\end{figure*}
\vspace{-0.5cm}
Sensors require fitting on different parts of the body and the number of sensors to be worn depends upon the context of the exercise. For instance, the Military Press exercise requires at least 3 IMUs for optimal performance. Despite their popularity, sensors may cause discomfort, thereby hindering the movement of participants. In addition, using multiple sensors leads to overheads such as synchronization, calibration and orientation.

Recent advances in computer vision have enabled the usage of 2D videos for human exercise classification. 
Past work explored posture detection \cite{Rahmadani} and the application of human exercise classification using pose estimation. Our previous work  \cite{ashishdami2022} proposed a novel method named BodyMTS to classify human exercises using video, human pose estimation and multivariate time series classification. There is less work comparing sensors with video in real-world applications. 
In this paper, we compare the performance of a sensor-based approach utilizing 5 IMUs with that of video from a single front-facing camera, on the same set of 54 participants, on two real-world datasets consisting of Military Press (MP) and Rowing exercises. These are important S\&C exercises and are widely used for injury risk screening and rehabilitation \cite{Darragh2016}. Incorrect executions may lead to musculoskeletal injuries and undermine the performance of athletes \cite{Baechle2008}. Hence, correct detection of abnormal movements is crucial to avoid injuries and maximize performance. 

The main requirements for an effective human exercise classification application are \cite{ashishdami2022}: accurate monitoring of body parts movement, correct classification of deviations from normal movements, timely feedback to end users, simple data capture using available smartphones and coverage of a wide range of S\&C exercises. Previous work \cite{whelan2019determining}  has shown that this task is difficult and has poor intra and inter-rater accuracy in user studies with domain experts, with Kappa scores for inter-rater agreement between  0.18-0.53, and intra-rater between 0.38-0.62. Through discussions with domain experts, we established that an effective application should achieve a minimum accuracy of 80\% to be useful for end users.

Existing methods using IMUs involve pre-processing the raw data, creating handcrafted features \cite{o2017classification, Martin2018}, and applying classical machine learning algorithms. Handcrafted feature extraction is often tedious and time-consuming, requires access to domain knowledge and is prone to cherry-pick features that only work for a specific set of exercises.
Deep learning methods \cite{Nutter2018} overcome this issue by automatically constructing features during training, but still require expertise in deep learning architectures along with hardware resources such as GPUs. Hence, we take two approaches to feature extraction: (1) using lightweight packages such as catch22 \cite{Lubba2019catch22CT} and tsfresh \cite{CHRIST201872} to automate the feature extraction from raw signals and (2) using the raw time series data with time series classifiers, which implicitly construct features inside the algorithm. 
For videos, we first extract multivariate data using human pose estimation with  OpenPose \cite{openpose2019} to obtain (X,Y) location coordinates of key body parts over all the frames of a video. 
Figure \ref{fig:overview} shows data captured with IMUs and video for the Military Press exercise. The top part shows the Y-signal for 3 body parts for a total of 10 repetitions, while the bottom part shows the X, Y, and Z signals of the magnetometer from an IMU worn on the right arm for the same set of 10 repetitions.  
\textbf{Our main contributions are:}
\begin{itemize}
    \item We compare 3 strategies for creating features from IMU data for human exercise classification. We observe that directly classifying the raw signals using multivariate time series classifiers outperforms the approach based on handcrafted features by a margin of 10 and 4 percentage points in accuracy for MP and Rowing respectively. Automatic feature extraction shows better performance than handcrafted features.
     
    \item We compare the performance of IMU and video for human exercise classification. We  observe that a single video-based approach outperforms a single IMU-based approach by a margin of 5 percentage points accuracy for MP and 15 percentage points for Rowing. 
    Additionally, we observe that a minimum of 3 IMU devices are needed to outperform a single video for both exercises.
    
    \item We propose an ensemble model that combines the data modalities from IMU and video, which outperforms either approach by a minimum of 2 percentage points accuracy for both MP and Rowing. This leads to an accuracy of 93\% for MP and 87\% for Rowing, using only a single IMU and a reduced-size video. We discuss reasons why combining video and sensor data is beneficial, in particular, the 2D video provides positional information, while the sensor provides information on orientation and depth of movement.
    \item To support this paper we have made all our code and data available  \footnote{\url{https://github.com/mlgig/Video_vs_Shimmer_ECML_2023}}. 
    % \footnote{\url{https://drive.google.com/drive/folders/1IoX5-GoO9w6PP1juUKnsrl6Ymn-BjiEY?usp=sharing}}.
\end{itemize}

The rest of the paper is organized as follows. Section \ref{sec:relwork} presents an overview of related work, Section \ref{sec:data} describes the data collection procedure, Section \ref{sec:data_analysis} describes the data analysis and methodology for classification and Section \ref{sec:experiments} presents the classification results using IMUs and video. Section \ref{sec:sig_impact} concludes and outlines directions for future work and Section \ref{sec:eth_impact} discusses ethical implications of this work.

%% file: 2-related-work.tex
\section{Related Work}
\label{sec:relwork}

This section describes the purpose of S\&C exercises and provides an overview of sensor-based and video-based data capture approaches. 

\subsection{S\&C Exercise Classification}

S\&C exercises aim at improving the performance of human participants in terms of strength, speed and agility, and they can be captured using sensor-based or video-based techniques.

Wearable sensor-based approaches involve fitting Inertial Measurement Units (IMUs) \cite{Martin2018, Martin2015} on different parts of the body. This is followed by creating handcrafted features which are used in conjunction with a classical machine learning model. Deep learning methods attempt to automate the process of feature extraction. CNN models work by stacking IMU signals into an image \cite{Nutter2018}, whereas \cite{Wenjin2021} uses an attention mechanism to identify the important parts in a signal.
Using IMUs has its own limitations. First, the number of inertial sensors required and their positions can vary from exercise to exercise \cite{Darragh2016, Martin2018, o2017classification}. Furthermore, sensors require calibration and synchronization and may also hinder the movement of the body and cause discomfort when used over longer time periods \cite{Darragh2016, Kwon2020}. 

Video-based systems can be categorized into 3 types: 3D motion capture, depth camera-based and 2D video camera. Though they are accurate, 3D motion capture systems are expensive and require complex setups. In addition, fitting multiple markers on the body may hinder the normal movement of the body \cite{Martin2018}. Microsoft Kinect is commonly used for depth camera-based systems \cite{zerpa2015use, ressman2020reliability, decroos0BVD18}. These systems are less accurate and are affected by poor lighting, occlusion, and clothing, and require high maintenance \cite{Martin2018}. The third subcategory uses video-based devices such as DSLR or smartphone cameras. Works based on video rely on human pose estimation to track different body parts \cite{Nobuyasu2020, ashishdami2022, ashish2020} and have shown 2D videos to be a potential alternative to IMU sensors. 
The video-based analysis also includes commercial software such as Dartfish  \cite{faro2016use} by providing the option to analyze motion at a very low frame rate. However, these are less accurate and require fitting body markers of a different colour to the background. 

\subsection{Multivariate Time Series Classification (MTSC)}

In multivariate time series classification tasks, the data is ordered and each sample has more than one dimension. 
We focus on recent linear classifiers and deep learning methods, which have been shown to achieve high accuracy with minimal run-time and memory requirements  \cite{Tan2021MultiRocket, ruiz2020great}.

\noindent\textbf{Linear Classifiers}. ROCKET \cite{Dempster2020} is a state-of-the-art algorithm for MTSC in terms of accuracy and scalability. Two more extensions named MiniROCKET \cite{Dempster2021MINIROCKETAV} and MultiROCKET \cite{Tan2021MultiRocket}, have further improved this method. These classifiers work by using a large number of random convolutional kernels which capture different characteristics of a signal and hence do not require learning the kernel weights as opposed to deep learning methods. These features are then classified using a linear classifier such as Logistic or Ridge Regression.

\noindent\textbf{Deep Learning Classifiers.} Deep learning architectures based on Fully Convolutional Networks (FCN) and Resnet \cite{Fawaz2019,ruiz2020great} have shown competitive performance for MTSC, without suffering from high time and memory complexity.

%% file: 3-data.tex
\section{Data Collection}
\label{sec:data}
    
    \textbf{Participants.}  
    54 healthy volunteers (32 males and 22 females, age: 26 $\pm$ 5 years, height: 1.73 $\pm$ 0.09 m, body mass: 72 $\pm$ 15 kg) were recruited for the study. Participants were asked to complete multiple repetitions of the two exercises in this study; the Military Press and Rowing exercises.  In each case, the exercises were performed under 'normal' and 'induced' conditions. In the 'normal' condition the exercise was performed with the correct biomechanical form and in the 'induced' condition the exercise was purposefully performed with pre-determined deviations from the normal form, assessed and confirmed in real-time by the movement scientist. Please refer to these sources \cite{ashish2020, ashishdami2022} for additional information on the experiment protocol.

    The data was collected using two video cameras and 5 Shimmer IMUs placed on 5 different parts of the body. Two cameras (30 frames/sec with 720p resolution) were set up in front and to the side of the participants. In this work, we only use the video recordings from the front view camera which is a more common use case. The 5 IMUs with settings: sampling frequency of 51.2 Hz, tri-axial accelerometer(±2 g), gyroscope (±500 $^{\circ}$/s) and magnetometer (±1.9 Ga) \cite{o2017classification} were fitted on the participants at the following five locations: Left Wrist (LW),  Right Wrist (RW),  Left Arm (LA),  Right Arm (RA) and Back. The orientation and locations of all the IMUs were consistent for all the participants. 
	
    \noindent\textbf{Exercise Technique and Deviations.}
    The induced forms were further sub-categorized depending on the exercise. 
    
    \subsection{Exercise Classes for Military Press (MP)} 
\begin{figure}[ht]
    \centering
    \includegraphics[scale=0.8, width=1.0\linewidth]{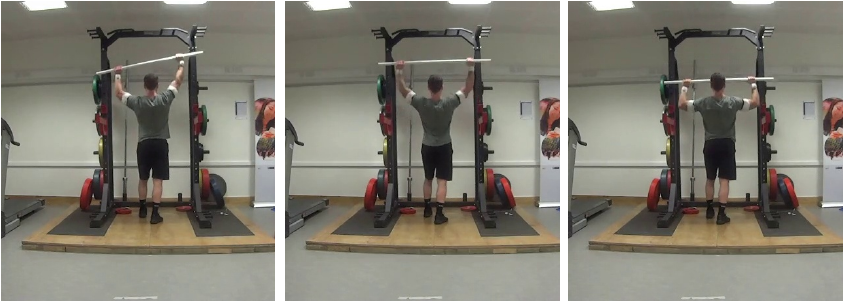}
    \caption{Single frames from the Military Press exercise, depicting the induced deviations for class A, Arch and R (left to right).}
    \label{fig:mp_deviations}
    \end{figure}

    \textbf{Normal (N):} This class refers to the correct execution, involving lifting the bar from shoulder level to above the head, fully extending the arms, and returning it back to shoulder level with no arch in the back. The bar must be stable and parallel to the ground throughout the execution.
    \textbf{Asymmetrical (A):} The bar is lopsided and asymmetrical.
    \textbf{Reduced Range (R):} The bar is not brought down completely to the shoulder level.
    \textbf{Arch (Arch):} The participant arches their back during execution.
    Figure \ref{fig:mp_deviations} shows these deviations using a single frame.
    \begin{figure}[ht]
    \centering
    \includegraphics[scale=0.9, width=1.0\linewidth]{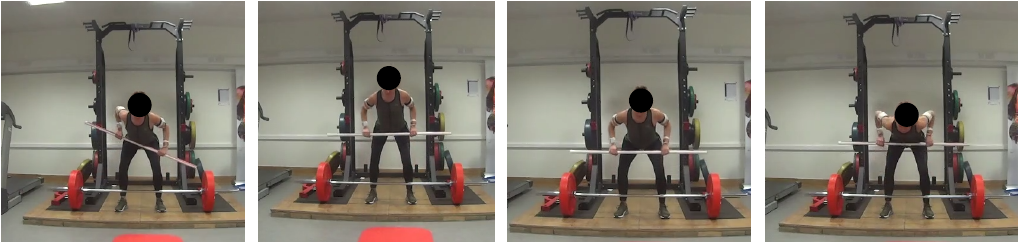}
    \caption{Single frames for the Rowing exercise, depicting the induced  deviations for class A, Ext, R and RB (left to right).}
    \label{fig:rowing_deviations}
    \end{figure}

    \subsection{Exercise Classes for Rowing} 
    
    \textbf{Normal (N):} This class refers to the correct execution, where the participant begins by positioning themselves correctly, bending knees and leaning forward from the waist. The execution starts by lifting the bar with fully extended arms until it touches the sternum and bringing it back to the starting position. The bar must be stable and parallel to the ground and the back should be straight.
    \textbf{Asymmetrical (A):} The bar is lopsided and asymmetrical.
    \textbf{Reduced Range (R):} The bar is not brought up completely until it touches the sternum.
    \textbf{Ext:} The participant moves his/her back during execution. 
    \textbf{RB:} The participant executes with a rounded back. 
    Figure \ref{fig:rowing_deviations} shows these deviations by depicting a single frame.

%% file: 4-methods.tex
\section{Data Analysis and Methods}
    \label{sec:data_analysis}
    This section presents the data pre-processing, features extraction and classification models. We present the feature extraction for IMU data, followed by feature extraction for video. We also provide a description of the train/test splits for IMUs and video data.

    \subsection{IMU Data}
    We discuss three strategies to create features from IMU data. First, we directly use the raw signal as a time series. Second, we use existing approaches to create handcrafted features. Third, we use dedicated packages to automatically extract features. Features extraction is performed after segmenting the full signal to obtain individual repetitions. 

    \subsubsection{Raw Signal as Multivariate Time Series.}
    The raw signal from IMU records data for 10 repetitions. Hence, we segment the time series to obtain signals for individual repetitions. The Y signal of the magnetometer from the IMU placed on the right arm is utilized to segment the signals. The time series obtained after this step has variable length since the time taken to complete each repetition differs from participant to participant. Further, current implementations of selected time series classifiers cannot handle variable-length time series and therefore all time series are re-sampled to a length of 161 (the length of the longest time series). This does not impact the performance as shown in the supplementary material.
    Every single repetition constitutes a single sample for train/test data. The final data D has a shape of $D \in \mathbb{R}^{N \times 45 \times 161}$, where N indicates the total samples.
    Each sample denoted by $x_{i}$ in the data has a dimension of $x_{i}$ $\in$ $\mathbb{R}^{45 \times 161}$, where 45 denotes the total number of time series (5 IMUs x 9 signals) and 161 is the length of each time series.
    
    \subsubsection{Handcrafted Features.}
    Each of the 5 IMUs outputs 9 signals $(X,Y,Z)$ for each of the accelerometer, magnetometer and gyroscope. 
    We follow the procedure as described in  \cite{o2017classification} to create handcrafted features. Additionally, 5 signals were created for each IMU: pitch, roll, yaw signal and vector magnitude of  accelerometer and gyroscope, 
    giving a total of 70 signals $(5 \times (9 + 5))$. For each repetition signal, 18 handcrafted features that capture time and frequency domain characteristics were created.
    Hence, we obtain the final data $D \in \mathbb{R}^{N \times 1260}$, where N is the total samples and 1260 represents the features extracted from 70 signals with 18 features each for both MP and Rowing.
    
    \subsubsection{Auto Extracted Features.}
    We use packages catch22 \cite{Lubba2019catch22CT} and tsfresh \cite{CHRIST201872} to perform automatic feature extraction from a single repetition signal. These packages calculate a wide range of pre-defined metrics in order to capture the diverse characteristics of a signal. They are straightforward to use and avoid the need for domain knowledge and signal processing techniques. 
    Catch22 captures 22 features for each of the 45 signals (5 IMUs x 9 signals) giving a total of 990 tabular features for MP and Rowing in the final dataset $D \in \mathbb{R}^{N \times 990}$, where N indicates the total samples. Similarly, tsfresh captures a large number of time series characteristics by creating a large number of features. 
    The final dataset D has a shape of $D \in \mathbb{R}^{N \times 15000}$ and $D \in \mathbb{R}^{N \times 16000}$, for MP and Rowing respectively. Both manual and automatic feature extraction are performed on the normalized time series, as we observed that normalizing the time series leads to an increase in accuracy.

    \subsection{Video Data} 
    We follow the methodology presented in our previous work \cite{ashishdami2022} to classify human exercise from videos. OpenPose is used for human pose estimation to track the key body parts, followed by a multivariate time series classifier. Each video consists of a sequence of frames where each frame is considered a time step. Each frame is fed to OpenPose which outputs coordinates $(X,Y)$ for 25 body parts. We only use the 8 upper body parts most relevant to the target exercises but also conduct experiments with the full 25 body parts.
    The time series obtained from a single body part is denoted by $b^n$ = $[(X,Y)^1, (X,Y)^2, (X,Y)^3,...(X,Y)^T]$ where n indicates the $n^{th}$ body part and T is the length of the video clip.

    \subsubsection{Multivariate Time Series Data.}
    Since each video records 10 repetitions for each exercise execution, segmentation is necessary in order to obtain single repetitions. Each repetition forms a single time series sample for training and evaluating a classifier. We use peak detection to segment the time series as mentioned in our previous work \cite{ashishdami2022}.
    Similarly to the IMU case, every time series obtained after this step has a variable length and therefore is re-sampled to a length of 161. 
    The final data is denoted by $D \in \mathbb{R}^{N \times 16 \times 161}$, where N indicates the total samples. Each sample denoted by $x_{i}$ has a dimension of $x_{i} \in \mathbb{R}^{16 \times 161}$, where 16 indicates $X$ and $Y$ coordinates for 8 body parts and 161 is the length of each time series.

    \subsubsection{Auto Extracted Features.}
    We use catch22 \cite{Lubba2019catch22CT} and tsfresh \cite{CHRIST201872} to perform automatic feature extraction from each single repetition signal.

    \vspace{-0.2cm}

    \subsection{Train/Test Splits}
    We use 3 train/test splits in the ratio of 70/30 on the full data set to obtain train and test data for both IMUs and video. Each split is done based on the unique participant IDs to avoid leaking information into the test data. 
    Train data is further split in the ratio of 85/15 to create validation data to fine-tune the hyperparameters. The validation data is merged back into the train data before the final classification.
    The data is balanced across all the classes. Table \ref{table:train_test_split_stats} shows the number of samples across all classes for a single train/test split for MP and Rowing respectively.

\begin{table}[!htb]
\caption{Samples per class in train/test dataset for a single 70/30 split for MP (left) and Rowing (right) for both IMU and video.}
\label{table:train_test_split_stats}
\begin{minipage}{.5\linewidth}
    \centering
 \begin{tabular}{llll}
     \toprule
        \textbf{Class} & \textbf{Train} & \textbf{Test} & \textbf{Total} \\
        \midrule
        N & 370 & 150 & 520 \\
        A & 340 & 150 & 490 \\
        R & 366 & 155 & 521 \\
        Arch & 350 & 140 & 490 \\
        \bottomrule
        \textbf{Total} & \textbf{1426} & \textbf{595} & \textbf{2021} \\
    \end{tabular}
\end{minipage}\hfill
\begin{minipage}{.5\linewidth}
    \centering

\begin{tabular}{llll}
     \toprule
        \textbf{Class} & \textbf{Train} & \textbf{Test} & \textbf{Total} \\
        \midrule
        N & 360 & 160 & 520 \\
        A & 362 & 150 & 512 \\
        Ext & 340 & 130 & 470 \\
        R & 380 & 150 & 530 \\
        RB & 361 & 140 & 501 \\
        \bottomrule
        \textbf{Total} & \textbf{1803} & \textbf{730} & \textbf{2533} \\
    \end{tabular}
\end{minipage}
\end{table}
\vspace{-0.75cm}

    \subsection{Classification Models}
    We use tabular machine learning models to work with handcrafted and automated features. Informed by previous literature on feature extraction for IMU data \cite{o2017classification, Martin2018}, we focus on Logistic Regression, Ridge Regression, Naive Bayes, Random Forest and SVM as classifiers for tabular data.
    We select ROCKET, MultiROCKET and deep learning models FCN and Resnet as recent accurate and fast multivariate time series classifiers \cite{bagnall2016great}.

%% file: 5-experiments.tex
\section{Empirical Evaluation}
\label{sec:experiments}

%This section presents the results of IMU and video data for human exercise classification on two popular S\&C exercises: MP and Rowing. It starts by presenting results on IMU data, followed by results on video data. Lastly, we present an ensemble model which combines both IMU and video data modalities.

% It is subdivided into three sections. The first section presents the classification results using handcrafted features, automated features and raw data directly from IMUs. We compare these results to select the best way of creating features from IMUs. We also present the results using combinations of IMUs placed on different parts of the body when using the raw data in order to identify the required minimum number of sensors.  The second section presents the classification results using video. We also present results using automated features from videos and compare these with the results using IMU data. The last section presents an ensemble model which combines both the IMUs and video modalities for human exercise classification. We present the above results 
We present results on IMU data, video data and combinations using ensembles.
We report average accuracy over 3 train/test splits for all the results. We use the \textit{sklearn} library \cite{sklearn} to classify tabular data and \textit{sktime} \cite{sktime}  to classify time series data. All the experiments are performed using Python on an Ubuntu 18.04 system (16GB RAM, Intel i7-4790 CPU @ 3.60GHz). 
The Supplementary Material \footnote{\url{https://github.com/mlgig/Video_vs_Shimmer_ECML_2023/blob/master/Supplementary_material.pdf}} presents further detailed results on 
leave-one-participant-out cross-validation, demographic results, execution time, as well as the impact of normalization and re-sampling length on the classification accuracy.

\subsection{Accuracy using IMUs}

We present the classification results using 3 different strategies for creating features from IMU data. For tabular features, we perform  feature selection to reduce overfitting and execution time. We use Lasso Regression \emph{(C=0.01)} with L1 penalty for feature selection, where C is the regularization parameter.
% Other dimensionality reduction techniques such as PCA, SVD, ExtraTreeClassifier and Chi Square Test, did not work well without impacting the accuracy. We have also evaluated several classifiers (e.g., ) and have found that Logistic Regression worked best among those.
Logistic Regression achieves the best performance followed by Ridge Regression and SVM. These results suggest that linear classifiers are best suited for this problem. Hence we only present results using Logistic Regression here.
We tune hyperparameters, particularly regularization parameter \emph{C} of Logistic Regression using cross validation. We observed that Logistic Regression (LR) with \emph{C=0.01} achieves the highest accuracy (Table \ref{table:imu_diff_features} presents results with Logistic Regression). 

% Table \ref{table:imu_diff_features} presents the average accuracy using manually crafted and automated features over 3 splits for MP and Rowing.

\begin{table}[h!]
 \caption{Average accuracy on test data over 3 splits for selected multivariate time series classifiers using IMU raw data as time series.}
    \centering
    \resizebox{0.4\columnwidth}{!}{%
    \begin{tabular}{lll}
    \toprule
        \textbf{Classifier} & \textbf{Acc MP} & \textbf{Acc Rowing} \\ 
        \midrule
        FCN & 0.86  & 0.77  \\  
        ResNet & 0.87  & 0.74 \\
        ROCKET & \textbf{0.91}  & \textbf{0.80}  \\
        MultiROCKET & \textbf{0.91}  & \textbf{0.81} \\
        \bottomrule
    \end{tabular}
    }
   
    \label{table:diff_classifiers}
    \end{table}

Table \ref{table:diff_classifiers} presents the results using raw data and multivariate time series classifiers. ROCKET achieves the best performance with MultiROCKET having similar accuracy for this problem. ROCKET has the added benefit that it can also work with unnormalised data and it is faster during training and prediction, so we select this classifier for the rest of the analysis.
%It has also shown to achieve the best performance on video data \citep{ashishdami2022}. 
%Henceforth, we use ROCKET to present results for raw time series data. 
% We tune hyperparameters of ROCKET except \emph{normalization} and observe no impact on the accuracy. The normalization flag is turned on for raw IMU data as setting it off leads to a 3 percentage points drop in accuracy. 
% We think one of the reasons for ROCKET performing better than the alternatives is that in ROCKET the user can de-activate the data normalisation step which is crucial for this application. For the other algorithms data normalisation is done explicitly or implicitly by the algorithm.
We analyse the average accuracy using all 5 IMUs as well as combinations of IMUs using raw time series with ROCKET as classifier. The goal is to select the minimum number of IMUs needed to achieve the best performance for MP and Rowing. Table \ref{table:imu_diff_features} presents the average accuracy over 3 splits obtained using all IMUs whereas Table \ref{table:imu_diff_body_parts} presents the average accuracy using different combinations of IMUs. 

\begin{table}[h]
 \caption{Average accuracy obtained on 5 IMUs data by using three feature selection strategies. Logistic Regression (LR) is used for tabular data, whereas ROCKET is used for time series classification.}
        \label{table:imu_diff_features}
    \centering
        \resizebox{0.5\columnwidth}{!}{%
        \begin{tabular}{lcc}
        \toprule
        \textbf{Feature Type} & \textbf{Acc MP} & \textbf{Acc Rowing} \\ 
        \midrule
        \textbf{Tabular} & & \\
        Handcrafted & 0.80 & 0.76 \\
        Automated (catch22) & 0.84 & 0.75 \\
        Automated (tsfresh) & 0.88 & 0.80 \\
        \textbf{Raw Signals} & & \\
        Time series & \textbf{0.91} & \textbf{0.80} \\
        \bottomrule
        \end{tabular}
        }
       
 %    \end{table}
    
 % \begin{table}[h!]
 \vspace{0.5cm}
  \caption{Average accuracy obtained using the different placement of IMUs over three train/test splits using raw data as time series with ROCKET as classifier. }
        \label{table:imu_diff_body_parts}
    \centering
        \resizebox{0.6\columnwidth}{!}{%
        \begin{tabular}{lcc}
        \toprule
        \textbf{Placement of IMU} & \textbf{Acc  MP} & \textbf{Acc Rowing} \\ 
        \midrule
        5 IMUs         & \textbf{0.91} & \textbf{0.80} \\
        RightWrist                      & 0.83 & 0.68 \\
        LeftWrist                       & 0.84 & 0.70 \\
        RightArm                       & 0.77 & 0.65 \\
        LeftArm                         & 0.76 & 0.66 \\
        Back                             & 0.71 & 0.71 \\
        LeftWrist + RightWrist             & 0.88 & 0.75 \\
        LeftWrist + RightWrist + Back        & \textbf{0.91} & \textbf{0.80} \\
        LeftArm + RightArm               & 0.82 & 0.70 \\
        LeftArm + RightArm + Back         & 0.86 & 0.78 \\
        
        \bottomrule
        \end{tabular}
        }
    \end{table}

\vspace{0.3cm}
    
\noindent\textbf{Results and Discussion}: 
From Table \ref{table:imu_diff_features} we observe that using raw data with ROCKET achieves the highest accuracy when compared to the approaches based on handcrafted and automated feature extraction. We tune hyperparameters of ROCKET using the validation data, particularly the \emph{number-of-kernels} and observe no impact on the accuracy. The normalization flag is set to True here as turning it off leads to a 4 percentage points drop in the accuracy. ROCKET can easily be run on a single CPU machine without the need for much engineering effort (only 2 parameters to tune) and dedicated hardware. It is much faster than using tsfresh or catch22 for feature extraction followed by classification.
% We note that more exercises need to be included to check for the generalizability of using raw data for exercise classification, but this 2 types of exercises show promising results.
% Additionally, automated feature extraction leads to an explosion of the feature space (15,000 for MP and 16,000 for Rowing) which risks overfitting the model on the data. 
Table \ref{table:imu_diff_body_parts} presents the accuracy using different combinations of IMUs placed on different parts of the body. 
Accuracy is lowest when using only a single sensor. Accuracy starts to increase as more IMUs are included, for both MP and Rowing. We observe that placing 1 IMU on each wrist and 1 at the back achieved the same accuracy as using all 5 IMUs. The accuracy jumps from 0.83 to 0.88 moving from one IMU placed on the right wrist to two IMUs placed on both wrists and finally jumps to 0.91 when adding one more IMU at the back for MP. Similar behaviour is observed for Rowing. This suggests that 3 IMUs are sufficient for these exercises. 
%Further, going from 2 IMUs placed on both wrists to 3 IMUs including the back plays a crucial role in improving the accuracy, since, without it, accuracy drops by 3 and 5 percentage points for MP and Rowing respectively. 
% The accuracy achieved using the same set of IMUs for MP is higher than the what is achieved for Rowing exercise. 
% Furthermore, accuracy achieved using a single IMU placed on either wrist is higher than what is achieved using IMUs placed on either of the arms.
  
\subsection{Accuracy using Video}
Here we present the results of classification using video as the data source. We report the average accuracy over 3 train/test splits for MP and Rowing. We also present results using tabular classifiers with automated features for comparison with the IMU based approach. For the raw data approach, we study the accuracy when involving different body parts, e.g., all 25, the 8 upper body parts suggested by domain experts and results using automated channel selection technique \cite{Dhariyal2022}. The normalization flag is set to False here as turning it on leads to a 4 percentage points drop in accuracy. This is in contrast to the setting configured for IMUs. We tune hyperparameters of ROCKET, particularly the \emph{number-of-kernels} and observe no impact on the accuracy. Table \ref{table:bodymts_accuracy} presents the average accuracy using these different approaches for classifying MP and Rowing exercises.
 \begin{table}[h!]
  \caption{Average accuracy obtained by ROCKET using video as data source for MP and Rowing over three train/test splits. }
\label{table:bodymts_accuracy}
    \centering
        \resizebox{0.5\columnwidth}{!}{%
        \begin{tabular}{lcc}
        \toprule
        \textbf{Feature Type} & \textbf{Acc MP} & \textbf{Acc Rowing} \\ 
        \midrule
        \textbf{Tabular} & & \\
        Automated (catch22)  & 0.69 & 0.70 \\
        Automated  (tsfresh) & 0.77 & 0.77 \\
        \textbf{Raw Signals} & & \\
        25 body parts           & 0.82 & 0.79 \\
        8 body parts            & \textbf{0.88} & \textbf{0.83} \\
        Elbow Pair \cite{Dhariyal2022}           & 0.83 & 0.82 \\
        \bottomrule
        \end{tabular}
        }
       
    \end{table}

\noindent\textbf{Results and Discussion}: 
From Table \ref{table:bodymts_accuracy} we observe that the average accuracy achieved using raw time series is highest when using the 8 body parts suggested by domain experts. Using automated features does not seem to work very well, in this case, achieving accuracy below 80\% for both exercises. Moreover, using channel selection techniques leads to an improvement by 1 and 3 percentage points in accuracy versus using the full 25 body parts.

\subsection{IMU versus Video}
We compare IMU and video data for human exercise classification, using the raw data approach for both IMU and video as it achieves the best performance. We report the accuracy, 
%We present the confusion matrix from MP exercise. We also present 
the execution time and the storage space required.
\begin{table}[h!] 
\caption{Comparison of accuracy obtained using IMUs and video for MP and Rowing.} %over three train/test splits. }
\label{table:summary_comparison}
\vspace{0.01cm}
\large
    \centering
        \resizebox{0.7\columnwidth}{!}{%
        \begin{tabular}{lcc}
        \toprule
        \textbf{Data Source} & \textbf{Acc MP} & \textbf{Acc Rowing} \\ 
        \midrule
        \textbf{Placement of IMUs} & & \\
        3 IMUs (Wrists and Back) &   0.91 & 0.80 \\
        1 IMU (LeftWrist)                      & 0.84 & 0.70 \\
        % \midrule
        \textbf{Video}  & & \\
        25 body parts           & 0.82 & 0.79 \\
        8 body parts            & 0.88 & 0.83 \\
        % \midrule
        \textbf{Ensemble: video and IMUs}  & & \\
        Video (8 body parts) + 3 IMUs               & \textbf{0.93} & \textbf{0.88} \\
        %Video (8 body parts) + 1 IMU RightWrist & \textbf{0.93} & 0.85 \\
        Video (8 body parts) + 1 IMU LeftWrist & \textbf{0.93} & 0.87 \\
        \bottomrule
        \end{tabular}
        }
        
%     \end{table}
% \begin{table}[h!]
\vspace{0.5cm}
 \caption{Average train/test time (minutes) obtained using IMUs and video as data sources for MP over three train/test splits. We also report the average test (i.e., prediction) time over a clip of 10 repetitions. }
\label{table:execution_time}
\vspace{0.2cm}
\large
    \centering
        \resizebox{0.8\columnwidth}{!}{%
        \begin{tabular}{lcc}
        \toprule
        \textbf{Data Type} & \textbf{Training Time} & \textbf{Test Time/} \\ 
         & (minutes)& \textbf{Test time per clip of 10 reps} \\
        \midrule
        \textbf{Sensor}                                    &     & \\
        3 IMUs  (Wrists and Back)                          & 8   & 6/0.10 \\
        \textbf{Video}                                     &     & \\
        8 body parts                                       & 52  & 22/0.37  \\
        \textbf{Ensemble: video and IMUs}                  &     & \\
        Video (8 body parts) + 5 IMUs                      & 60  & 29/0.50 \\
        Video (8 body parts) + 1 IMU                       & 58  & 27/0.46  \\
        \bottomrule
        \end{tabular}
        }
       
    % \end{table}
%  \textbf{Sensor} & & \\
%         3 IMUs  (Wrists and Back)                          & 5/3 & 6/3 \\
%         \textbf{Video}  & & \\
%         8 body parts            & 52/22 & 57/24 \\
%         \textbf{Ensemble: video and IMUs}  & & \\
%         Video (8 body parts) + 5 IMUs               & 57/25 & 63/27 \\
%         Video (8 body parts) + 1 IMU               & 55/24 & 61/25 \\
 % \begin{table}[h!]
 \vspace{0.5cm}
  \caption{Storage consumption using raw videos, IMUs and video as time series for MP and Rowing exercises for the 54 participants in our study.}
    \label{table:data_size}
    \centering
    \resizebox{0.4\columnwidth}{!}{%
    \begin{tabular}{lll}
        \toprule
        \textbf{Data Size (MB)} & \textbf{MP} & \textbf{Rowing} \\
        \hline
        5 IMUs & 640 & 591\\
        Raw Videos (720p) & 813  & 1012 \\
        Videos as Time Series & 97  & 114  \\ \hline
    \end{tabular}
    }   
    \end{table}
%\vspace{-0.5cm}
Table \ref{table:summary_comparison} presents the results for both MP and Rowing exercises. We observe that a minimum of 3 IMUs are required to achieve a higher accuracy than a single video.
A single video outperforms a single IMU for both exercises by a minimum of 5 percentage points.
Table \ref{table:execution_time} reports the real train/test time for both approaches. This time includes  time taken for data pre-processing and to train/test the model. It also includes time to run pose estimation in case of video.
The IMUs approach takes the least amount of time to train/test as compared to the video-based approach. For video, OpenPose  extracts the multivariate time series data. The total duration of all videos is 1h 38 minutes for MP, whereas OpenPose took 1h 12 minutes thus OpenPose can run faster than real-time, which is important for getting fast predictions.
Table \ref{table:data_size} presents the storage consumption for both approaches. We note savings in terms of storage space:  5 IMUs require 6 times more space than the time series obtained from videos. Even after selecting the minimum number of sensors which is 3 in both exercises, the storage consumption is more than 200 MB which is also higher as compared to using time series from video.
Our previous work in \cite{ashishdami2022} explored the impact of video quality such as resolution and bit rate on classification accuracy and demonstrated how much video quality can be degraded without having a significant impact on the accuracy, whilst saving storage space and processing power.  
 
\subsection{Combining IMU and Video}
We create an ensemble model by combining individual models trained independently on IMU and Video. For IMUs, we take the 3 sensors that achieved the highest accuracy. When video is combined with just a single sensor, we take the IMU placed on the left wrist, as it had the highest accuracy among single sensors and it is the most common location for people to wear their smartwatch. 
%is a more commonly used configuration.  
Probabilities are combined by averaging and the class with the highest average probability is predicted for a sample during test time. Table \ref{table:summary_comparison} presents a comparison of different approaches, using ROCKET as a multivariate time series classifier. From Table \ref{table:summary_comparison}, we observe that an ensemble model achieves the best average accuracy when compared to using any number of IMUs and a single video-based approach. The accuracy for MP jumps by 2 percentage points when transitioning from 5 IMUs to an ensemble approach, and by 5 percentage points when moving from a single video to an ensemble. Similar results are observed for Rowing. These results suggest that combining IMU and video modalities enhances the performance of exercise classification. Combining video and IMU data sources, with video providing 2D location coordinates for key anatomical landmarks and IMUs capturing acceleration and orientation of the body parts, results in improved classification accuracy, as shown in this investigation (see supplementary material). This finding is consistent with previous work in \cite{Marcard2016HumanPE} that highlights the complementary nature of video and IMUs in enhancing human pose estimation quality, while in this work we see a similar benefit for human exercise classification.

%% file: 6-conclusion.tex
\section{Conclusion }
\label{sec:sig_impact}
We presented a comparison of IMU and video-based approaches for human exercise classification on two real-world S\&C exercises (Military Press and Rowing) involving 54 participants.
We compared different feature-creation strategies for classification. The results show that an automated feature extraction approach outperforms classification that is based on manually created features. Additionally, directly using the raw time series data with multivariate time series classifiers achieves the best performance for both IMU and video. While comparing IMU and video-based approaches, we observed that using a single video significantly outperforms the accuracy obtained using a single IMU. Moreover, the minimum number of IMUs required is not known in advance, for instance, 3 IMUs are required for MP to reach a reasonable  accuracy. 
Next, we compared the performance of an ensemble method combining both IMU and video with the standalone approaches. 
We showed that an ensemble approach outperforms either data modality deployed in isolation. The accuracy achieved was 93\% and 88\% for MP and Rowing respectively. 
% This is a very encouraging result that brings this application closer to practice.
The criteria to select sensors or videos will ultimately depend on the goal of the end user. For instance: the choice between video and IMUs will depend on a combination of factors such as convenience and levels of accuracy required for the specific application context.  

% While the above results show that video can be an alternative for the task of human exercise classification, further research is needed to generalize the above results by including more exercises, e.g., to cover both upper limbs and lower limbs exercises.

We acknowledge the fact that the scenario that was tested in this research does not accurately reflect real-world conditions. This does mean that we are exposed to the risk that the induced deviation performances could be exaggerated, and therefore not reflective of the often very minor deviations that can be observed in the real-world setting.  However, we would argue that performing exercises under induced deviation conditions, if done appropriately, is a very necessary first step towards validating these exercise classification strategies in this field. It would not be prudent to assume that this model could be generalised to operate to the same level in real-world conditions. Having said that, the use of conditioned datasets is a necessary first step in this kind of application and provides the proof of concept evidence necessary to move onto the real-world setting. 

\subsubsection{Acknowledgements} This work was funded by Science Foundation Ireland through the Insight Centre for Data Analytics (12/RC/2289\_P2) and VistaMilk SFI Research Centre
(SFI/16/RC/3835).

%% file: ethical_issues.tex
\section{Ethical Implications}
\label{sec:eth_impact}
%  https://nips.cc/public/EthicsGuidelines

% The previous results show that video can also serve as an alternative or complementary data to using sensors for the task of human exercise classification. 

Using videos for human exercise classification raises ethical implications that need to be mitigated, prompting a discussion of potential ethical implications.

\noindent\textbf{Data Collection.}
Participants in this study provided written consent and the Human Research Ethics Committee of the university approved this study. All experiments were conducted under the supervision of an expert physiotherapist. The potential implications, in this case, can arise when the language used for the consent form may not be native to all the participants. In our case, the organizing authority or professional who was carrying out the data collection made sure that all the participants have well understood the consent form and the use of this data in the future.

\noindent\textbf{Privacy and Confidentiality.} This study uses videos which record participants executing exercises. This poses obvious privacy challenges. A first step is to blur the video to protect the participant's identity. This work utilizes human pose estimation to extract time series from video, thereby avoiding the need to directly use the original video. By working with the extracted time series, it largely safeguards the privacy and confidentiality of the participants.

\noindent\textbf{Diversity of Representation.}
The participants considered in this study fall into the age group of 20 to 46. Hence the results presented here may not generalise for other age groups. Therefore the final use case will depend on the specific target users, such as athletes competing in the Olympic games versus individuals with less intensive training goals. While there were slightly more male participants than female participants, it does not impact the conclusions drawn in this work, as analysed in the supplementary material. However, this requires further exploration to avoid any biases in the conclusion. Future studies should aim for equal representation among participants in terms of age, sex, gender, race etc., from the start of the study. 

\noindent\textbf{Transparency and Feedback.} The prediction of the model in this case outputs whether the execution of the exercise was correct or incorrect. Deep learning-based models and other posthoc explanation methods support saliency maps which can be used to highlight the discriminative regions of the data that can be mapped back to the original video thus providing more information about the model decision to the participant. 

The above list is not exhaustive and other inherent biases may appear because of the chosen model and the way the data has been collected.